\documentclass{llncs}
\usepackage{epsf}
\usepackage{multirow}
\usepackage[flushleft]{threeparttable}
\usepackage{enumitem}
\usepackage[utf8]{inputenc}
\usepackage{graphicx}
\usepackage[usenames]{color}
\usepackage{ulem}
\usepackage{float}
\graphicspath{{images/}}

\begin{document}

\title{Friends and Enemies of Clinton and Trump: Using Context for Detecting Stance \\ in Political Tweets}

\author{
Mirko Lai\inst{1,2}, Delia Irazú Hernández Farías\inst{1,2}, Viviana Patti\inst{1} and Paolo Rosso\inst{2}}
\institute{Dipartimento di Informatica, Università degli Studi di Torino, Torino, Italy \and
Pattern Recognition and Human Language Technologies Research Center,
Universitat Politécnica de Valéncia, Valencia, Spain
}

\maketitle

\begin{abstract}

Stance detection, the task of identifying the speaker's opinion towards a particular target, has attracted the attention of researchers. 
This paper describes a novel approach for detecting stance in Twitter. 
We define a set of features in order to consider the context surrounding a target of interest with the final aim of
training a model for predicting the stance towards the mentioned targets.
In particular, we are interested in investigating political debates in social media. For this reason we evaluated our approach 
focusing on two targets of the SemEval-2016 Task 6 on Detecting stance in tweets,
which are related to the political campaign for the 2016 U.S. presidential elections: Hillary
Clinton vs. Donald Trump.
For the sake of comparison with the state of the art, we evaluated our model against the dataset released in the SemEval-2016 Task 6
shared task competition.
Our results outperform the best ones obtained by participating teams, and show that information about enemies and friends of politicians help in detecting stance towards them.
\end{abstract}

\section{Introduction}

Social media provide a way for expressing opinions about different topics.
From this kind of user-generated content it is possible to discover relevant information under several perspectives.
A wide range of research has been carried out in order to exploit the vast amount of data generated in social media.
One of the most interesting research areas concerns to investigate how people expose their feelings, evaluations, attitudes and emotions. These kinds of aspects are the subject of interest of Sentiment Analysis (SA) \cite{Liu-2012}. 

Determining the subjective value of a piece of text is the most general task of SA. 
Recently, the interest on studying finer-grained and different facets of sentiment in texts has derived in areas such as \textit{Aspect based sentiment analysis} \cite{pontiki-EtAl:2016:SemEval} and \textit{Stance Detection} (SD) \cite{mohammad-EtAl:2016:SemEval}, which is the focus of our work.
Identifying the speaker's opinion towards a particular target is the main goal of SD.
It is not enough to recognize whether or not a text is positive/negative/neutral but it is necessary to infer the point of view of the tweeter towards a particular target.

Stance detection could not only provide useful information for improving the performance of SA but it could also help to 
better understand the way in which people communicate ideas in order to highlight their point of view towards a particular 
target entity. This is particularly interesting when the target entity is controversial issue (e.g., political reforms \cite{BOSCO16.534,STRANISCI16.1063}) or a polarizing person (e.g., candidates in political elections). 
Therefore, detecting stance in social media could become a helpful tool for various sectors of society, such as 
journalism, companies and government, having politics as an especially good application domain. 
Several efforts have been made in order to investigate different aspects related to social media and politics \cite{Maldonado2016}. 
We are interested in political debates in social media, particularly in the interaction between polarized communities.
We consider that being able to detect stance in user-generated content could provide useful insights to discover novel information about social network structures.
Political debate texts coming from social media where people discuss their different points of view offer an attractive information source. 

This year, for the first time a shared task on stance detection in tweets was organized \cite{mohammad-EtAl:2016:SemEval}. Two of the targets considered in order to evaluate stance detection systems were: Hillary Clinton and Donald Trump\footnote{They are the candidates who won the Party Presidential Primaries for the Democratic and Republican parties, respectively.}.
Both targets have been the focus of different research, for instance in \cite{DBLP:journals/corr/SchumacherE16} the authors studied their speeches during the 2016 political campaign. 
In such way, studying these targets is an attracting topic of research due to the impact of the use of social media during the political campaign for the 2016 U.S. Presidential elections.

Our approach to detect stance in tweets relies mainly on the context of the targets of interest: Hillary Clinton and Donald Trump.
Besides, we also took advantage of widely used features in SA.

The paper is organized as follows. 
Section 2 introduces the first shared task on Twitter stance detection. 
Section 3 describes our method to detect stance by exploiting different features. 
Section 4 describes the evaluation and results. Finally, Section 5 draws some conclusions.

\section{Detecting Stance on Tweets} \label{sharedTask}

The SemEval-2016 Task 6: Detecting Stance in Tweets\footnote{\url{http://alt.qcri.org/semeval2016/task6/}} 
was the first shared task on detecting stance from tweets.
Mohammad et. al in \cite{mohammad-EtAl:2016:SemEval} describe the task as: \textit{Given a tweet text and a target entity (person, organization, movement, policy, etc.), automatic natural language systems must determine whether the tweeter is in favor of the target, against the given target, or whether inference is likely.}

Let us to introduce the following example\footnote{This tweet was extracted from the training set of SemEval-2016 Task 6.}: \\
\textit{Support \#independent \#BernieSanders because he's not a liar. \#POTUS \\ \#libcrib \#democrats \#tlot \#republicans \#WakeUpAmerica \#SemST} 

The target of interest is ``Hillary Clinton". Here, the tweeter expresses a positive opinion towards an adversary of the target. Consequently the annotator inferred that the tweeter expresses a negative opinion towards the target. 
As can be noticed, this tweet does not contain any explicit clue to find the target. 

For evaluating the task, the organizers annotated near to 5,000 English tweets for stance towards six commonly known targets in the United States: ``Atheism", ``Climate Change is a Real Concern", ``Feminism Movement", ``Hillary Clinton", ``Legalization of Abortion", and ``Donald Trump" (Stance Dataset, henceforth).
A set of hashtags widely used by people when tweeting about these targets was compiled; then it was used to retrieve tweets according three categories: in-favor hashtags, against hashtags and stance-ambiguous hashtags.
The tweets were manually annotated by crowdsourcing. More details about the Stance Dataset can be found in \cite{mohammad-EtAl:2016:SemEval}.

The participants in the SemEval-2016 Task 6 were required to classify tweet-target pairs into exactly one of three classes: \textit{Favor}: It can be inferred from the tweet that the tweeter supports the target (e.g., directly or indirectly by supporting someone/something, by opposing or criticizing someone/something opposed to the target, or by echoing the stance of somebody else); \textit{Against}: It can be inferred from the tweet that the tweeter is against the target (e.g., directly or indirectly by opposing or criticizing someone/something, by supporting someone/something opposed to the target, or by echoing the stance of somebody else); and \textit{Neither}: None of the above.

The SemEval-2016 Task 6 was divided into two subtasks: 
\begin{itemize}[leftmargin=*]
    \item Task A. Supervised Framework. The participating systems were asked to perform stance detection towards the following targets: ``Atheism", ``Climate Change is a Real Concern", ``Feminism Movement", ``Hillary Clinton", and ``Legalization of Abortion". For evaluation the organizer provided a  training (2,914 tweets) and test (1,249 tweets) sets. 
    \item Task B. Weakly Supervised Framework. The task was detecting stance towards one target ``Donald Trump" in 707 tweets. For this task the participants were not provided with any training data about this target. 
\end{itemize}

Nineteen teams participated in Task A while only nine competed in Task B.
It is important to highlight that only two systems were evaluated specifically on Task B. Figure 1 shows a brief summary of the systems. Further information about the systems in the task can be found in \cite{SemEval:2016}\footnote{Notice that not all the reports describing systems and approaches of teams participating at SemEval-2016 Task 6 are available in \cite{SemEval:2016}.}.

Both tasks were addressed in similar ways. 
Most teams exploited standard text classification features such as n-grams and word embedding vectors. 
Besides, some SA features from well-known lexical resources, such as \textit{EmoLex} \cite{Mohammad:13}, \textit{MPQA} \cite{Wilson:2005}, 
\textit{Hu and Liu} \cite{HuLiu:2004} and \textit{NRC Hashtag} \cite{MohammadKZ2013}, were used to detect stance in tweets. 
Furthermore, some teams decided to take advantage of additional data by harvesting Twitter using  stance-bearing hashtags in order to have more stance tweets. 
It is important to highlight that the best system in Task A (MITRE) did use this alternative.
A similar approach was adopted by the three best ranked systems on Task B (pkudblab, LitisMind, and INF-UFRGS). For what concerns to Task B, in order to deal with the lack of training data, some systems attempted to generalize the supervised data from task A in different ways such as defining rules or by exploiting multi-stage classifiers.

%

\begin{table}[]
\centering
\caption{Brief description of the participating systems at SemEval-2016 Task 6}
\label{relatedWork}
\begin{tabular}{|c|p{9.5cm}|}
\hline
 System & \ \ Description \\ \hline 
MITRE \cite{zarrella-marsh:2016:SemEval} & \textbf{\tiny Overall approach:}  Recurrent neural networks. \\
Task A  & \textbf{\tiny External resources:} Words embeddings with the word2vect skip-gram method. Near to 300,000 tweets containing hashtags related to the targets.  \\ \hline
pkudblab \cite{wei-EtAl:2016:SemEval}  & \textbf{\tiny Overall approach:} Convolutional neural network. \\
Tasks A and B & \textbf{\tiny External resources:} Words embeddings using the Google News dataset.  \\ \hline
TakeLab \cite{tutek-EtAl:2016:SemEval} & \textbf{\tiny Overall approach:} An ensamble of learning algorithms (such as SVM, random forest) fine-tuned using a genetic algorithm. \\
Task A & \textbf{\tiny External resources:}  Word features, word embeddings, frequency of emoticons, uppercase characters, among others. \\ \hline
ECNU \cite{zhang-lan:2016:SemEval} & \textbf{\tiny Overall approach:}  A pipeline-based procedure involving relevance and orientation detection. \\
Tasks A and B & \textbf{\tiny External resources:} N-grams, topic features and sentiment lexicon features (such as Hu\&Liu and MPQA, among others).  \\ \hline
CU-GWU \cite{elfardy-diab:2016:SemEval} &  \textbf{\tiny Overall approach:} Classification using SVM \\
Task A  & \textbf{\tiny External resources:} N-grams, Stanford's SA system and LIWC.  \\ \hline
IUCL-RF \cite{liu-EtAl:2016:SemEval}  & \textbf{\tiny Overall approach:} Classification algorithms (SVM, random forest, gradient boosting decision trees) and an ensamble classifier (TiMBL). \\
Task A  & \textbf{\tiny External resources:} Bag-of-Words and word vectors.  \\ \hline
DeepStance \cite{vijayaraghavan-EtAl:2016:SemEval} & \textbf{\tiny Overall approach:} A set of naive bayes classifiers using deep learning. \\ 
Task A & \textbf{\tiny External resources:} More than 1.5 million of tweets were added by using representative hashtag for target-stance pairs.  \\ \hline
UWB \cite{krejzl-steinberger:2016:SemEval} & \textbf{\tiny Overall approach:} Maximum entropy classifier. \\
Tasks A and B & \textbf{\tiny External resources:} N-grams, PoS labels, General Inquirer. Additional tweets were gathered based on frequent hashtags in the training set. \\ \hline
IDI@NTNU \cite{bohler-EtAl:2016:SemEval} & \textbf{\tiny Overall approach:} A soft voting classifier approach (naive bayes and logistic regression). \\
Task A & \textbf{\tiny External resources:}  Word vectors, n-grams, char-grams, negation, punctuation marks, elongated words, among others.  \\ \hline
Tohoku \cite{igarashi-EtAl:2016:SemEval} & \textbf{\tiny Overall approach:} Two methods: a feature based approach and a neural network based approach. \\
Task A  & \textbf{\tiny External resources:} Bag-of-Words, PoS labels, SentiWordNet. Additional Twitter data was gathered from target words.   \\ \hline
ltl.uni-due \cite{wojatzki-zesch:2016:SemEval} & \textbf{\tiny Overall approach:} Multidimensional classification problem \\
Tasks A and B & \textbf{\tiny External resources:} N-grams, punctuation marks, negation, nouns.   \\ \hline
JU\_NLP \cite{patra-das-bandyopadhyay:2016:SemEval} & \textbf{\tiny Overall approach:} Classification using SVM \\
Task A & \textbf{\tiny External resources:} N-Gram and sentiment analysis resources such as: SentiWordNet, EmoLex and NRC Hashtag Emotion Lexicon.  \\ \hline
nldsucsc \cite{misra-EtAl:2016:SemEval} & \textbf{\tiny Overall approach:}  Classification using SVM, J48 and naive bayes. \\
Task A  & \textbf{\tiny External resources:} N-grams, PoS labels, LIWC. Additional tweets were gathered based on frequent hashtags in the training set.  \\ \hline
INF\_UFRGS \cite{dias-becker:2016:SemEval} & \textbf{\tiny Overall approach:} Set of rules together with SVM. \\
Task B & \textbf{\tiny External resources:} N-grams.  \\ \hline
USFD \cite{augenstein-vlachos-bontcheva:2016:SemEval} & \textbf{\tiny Overall approach:} Classification using logistic regression. \\
Task B & \textbf{\tiny External resources:} Bag-of-words autoencoder. Additional tweets were gathered by using two keywords per target.  \\ \hline
\end{tabular}
\end{table}

\section{Our approach} \label{ourApproach}
We are proposing a supervised approach for stance detection \footnote{\url{https://github.com/mirkolai/Friends-and-Enemies-of-Clinton-and-Trump}}.
Our work is focused on detecting stance towards Hillary Clinton and Donald Trump that are currently contesting the political campaign for the 2016 U.S. Presidential election.
An important aspect to mention concerns to the fact that when the Stance Dataset was built the two targets were still participating to the Party Presidential Primaries for the Democratic and Republican parties, respectively. 
We address the stance detection in tweets, casting it as a classification task. 
A set of features that comprises different aspects was exploited.
The most novel one refers to the extraction of context-related information regarding to the target of interest.
Our hypothesis is that domain knowledge could provide useful information to improve the performance of SD systems.
For instance, in order to correctly identify stance in a tweet as the one mentioned in Section \ref{sharedTask}, it is needed to recognize that \textit{Bernie Sander} was an adversary of Hillary Clinton during the Party Presidential Primaries of the Democratic party. 
Attempting to capture information related to domain knowledge,
we define two concepts: ``enemies" and ``friends".
These concepts are used for denoting the entities related to the target.
By using the terms ``enemies" and ``friends", we are trying to infer that when a tweeter is against an ``enemy"/``friend" of the target, then the tweeter is in favor/against towards the target and, on the other hand, when a tweeter is in favor towards an ``enemy"/``friend" of the target, then the tweeter is against/in favor towards the target. Figure 1 shows an example of the relationships between the ``friends" and ``enemies" according to their political party, in this case the target of interest is Hillary Clinton. \\
Three groups of features were considered: sentiment, structural, and context-based.
\begin{figure}[H]  
 \center
 \includegraphics[width=10cm, height=3cm]{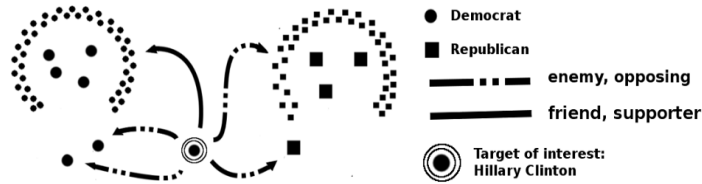} 
 \caption{Diagram of relationships between friends and enemies of Hillary Clinton}
\end{figure} 
%
\subsection*{Sentiment-based Features}

We shared the idea that stance detection is strongly related to sentiment analysis
\cite{mohammad-EtAl:2016:SemEval,zhang-lan:2016:SemEval}.
As far as we know, there are not sentiment analysis lexica retrieved specifically in the political domain\footnote{For example, the term \textit{vote} is strongly related to politics, but it is not present in commonly used SA lexica such as: AFINN, Hu\&Liu, and LIWC.}; thus, in order to take advantage of sentiment features it is possible to exploit the wide range of resources available for English.
We used a set of four lexica to cover different facets of affect ranging from prior polarity of words to fine-grained emotional information: 
\begin{itemize}[leftmargin=*]
\setlength\itemsep{1em}
\item \textbf{AFINN}. It is an affective lexicon of 2,477 English words manually labeled with a polarity value between -5 to +5. AFINN was collected by Finn Årup Nielsen \cite{Nielsen2011ANEW}. We consider one feature from AFINN: the sum of the polarity of the words present in each tweet. 
\item \textbf{Hu\&Liu (HL)}. It includes about 6,800 positive and negative words. We calculate the difference between the positive and negative words in a tweet as a feature. 
\item \textbf{LIWC}. The Linguistic Inquiry and Word Counts (LIWC) \cite{Pennebaker:01} is a dictionary that contains about 4,500 entries distributed in 64 categories that can be further used to analyse psycholinguistic features in texts. We calculate the difference between PosEmo (with 405 entries) and  NegEmo (with 500 entries) categories in a tweet as a feature.
\item \textbf{DAL}. The Dictionary of Affect in Language (DAL) contains 8,742 English words; it was developed by Whissell \cite{Whissell01102009}. Each word is rated in a three-point scale into three dimensions: Pleasantness (It refers to the degree of pleasure produced by words), Activation (It refers the degree of response that humans have under an emotional state) and Imagery (It refers to how difficult to form a mental picture of a given word is).
We consider six features, i.e. the sum and the mean of the rates of the words present in the tweet for each one of the three dimensions.
\end{itemize}

\subsection*{Structural Features}

We also explore structural characteristics of tweets because we believe that could be useful to detect stance. 
We experimented with several kinds of structural features, however only the most relevant ones were included in the final approach:
\begin{itemize}[leftmargin=*]
\setlength\itemsep{1em}
\item \textbf{Hashtags}. The frequency of hashtags present in each tweet.
\item \textbf{Mentions}. The frequency of screen names (often called mentions) in each tweet.
\item \textbf{Punctuation marks (punct\_marks)}. We consider a set of 6 different features: the frequency of exclamation marks, question marks, periods, commas, semicolons, and finally the sum of all the punctuation marks mentioned before.
\end{itemize}

\subsection*{Context-based Features}

Our hypothesis is that the context-based features should capture some domain-related information. 
An overall perspective of the context surrounding a target can be acquired by the relationships that exist between the target and other entities in its domain.
As mentioned before we are interested in investigating Political debates: for this reason we selected as targets of interest politicians such as Hillary Clinton and Donald Trump.
We manually created a list of entities related to the Party Presidential Primaries for the Democratic and Republican parties from Wikipedia\footnote{Articles: \textit{Democratic Party presidential primaries, 2016} and \textit{Republican Party presidential candidates, 2016}}.
We exploited 6 types of context-based features considering different kinds of relationships between the target and the entities around the target:

\begin{itemize}[leftmargin=*]
\setlength\itemsep{1em}
\item \textbf{Target of interest mentioned by name (targetByName)}: 
This feature captures the presence of the target of interest in the tweet in hand.
\textit{\#Stop\textbf{Hillary}2016 \@\textbf{HillaryClinton} if there was a woman with integrity and honesty I would vote for such as woman president, NO}. 
The list of tokens used to check the presence of the target of interest are: \textit{hillaryclinton}, \textit{hillary}, \textit{clinton}, and \textit{hill} for Hillary Clinton; while  for Donald Trump are \textit{realdonaldtrump}, \textit{donald}, and \textit{trump}.
\item \textbf{Target of interest mentioned by pronoun (targetByPronoun)}:
This feature allows to identify those cases when the target of interest is mentioned by using a pronoun. 
In the following example, knowing that the target of the tweet is Hillary Clinton, it is possible to exploit the pronoun ``she" to capture the presence of the target in hand.
\textit{\@HomeOfUncleSam \@ScotsFyre \@RWNutjob1 \@SA\_Hartdegen \textbf{She}'s too old to understand the internet...that \textbf{she} can be fact checked.} \\
Two pronouns were considered for each one of the targets of interest: \textit{she} and \textit{her} for Hillary Clinton, while \textit{he} and \textit{his} for Donald Trump. 
\item \textbf{Target's party (targetParty)}:
As people involved in politics, our targets belong to a political party. Using this feature we identify if the stance against (or in favor) towards the target of interest was expressed mentioning the name of the party instead of the target.
In the following example the tweeter expresses a negative opinion toward Hillary Clinton party. \\
\textit{It's a miracle, suddenly \#\textbf{Democrats} don't mind having someone who voted for war.} \\
In this  case we consider the tokens \textit{dem}, \textit{democratic}, \textit{democrat}, \textit{democrats}, \textit{progressive} in order to check the entity party for Hillary Clinton, while
we consider the tokens \textit{republican}, \textit{republicans}, and \textit{conservative} for Donald Trump.
\item \textbf{Party colleague opposite (targetPartyColleagues)}:
We also considered the case where the party colleagues of the target of interest are mentioned to express an opinion towards it. We use the name and the surnames of the candidates for the Party Presidential Primaries for both Democratic and Republican parties.
In the example, Hillary Clinton's party colleagues are mentioned.
\textit{\@msnbc \@\textbf{Lawrence} \@JoeBiden \@Sen\textbf{Sanders}  we love Joe and \textbf{Bernie}--but they ARE too OLD--they would end up a \#OneTerm President \#SemST}
\\
The list of names used for Hillary Clinton is: \textit{bernie}, \textit{sanders}, \textit{martin}, \textit{o'malley}, \textit{lincoln}, \textit{chafee}, \textit{webb}, \textit{lawrence}, and \textit{lessig}; while for Donald Trump is: \textit{ted}, \textit{cruz}, \textit{marco}, \textit{rubio}, \textit{john}, \textit{kasich}, \textit{ben}, \textit{carson}, \textit{jeb}, \textit{bush}, \textit{rand}, \textit{paul}, \textit{mike}, \textit{huckabee}, \textit{carly}, \textit{fiorina}, \textit{chris}, \textit{christie}, \textit{rick}, \textit{santorum}, \textit{gilmore}, \textit{rick}, \textit{perry}, \textit{scott}, \textit{walker}, \textit{bobby}, \textit{jindal}, \textit{lindsey}, \textit{graham}, \textit{george}, \textit{pataki}.
\item \textbf{Target's oppositors party (targetsOppositors)}:
This feature captures the presence of oppositors belonging to the rival party of target of interest's.
In the following example a positive opinion is expressed towards two candidates from the Republican party. Thus, the tweet is against Hillary Clinton.\\
\textit{\@PhilGlutting \@megadreamin Thank you so much for RT and FAV!!! \#WakeUpAmerica \#\textbf{Rubio}2016 \#\textbf{Cruz}2016 \#SemST}\\
We use the Donald Trump's tokens lists targetParty and targetPartyColleagues in order to create Hillary Clinton's targetsOppositors tokens list, while we use Hillary Clinton's tokens  lists targetParty and targetPartyColleagues in order to create Donald Trump's targetsOppositors tokens list.
\item \textbf{Nobody (nobody)}:
This feature allows to catch those cases where any of the above described entities are mentioned in a tweet.
In the following example the term \textit{Ambassador} refers to Chris Stevens, who served as the U.S. Ambassador to Libya and who was killed at Bengasi in 2012.
The diplomat is related to Hillary Clinton in a situation not related with the election campaign\footnote{\url{https://en.wikipedia.org/wiki/J.\_Christopher\_Stevens}}.\\
\textit{I don't want to be appointed to an Ambassador post.}\\The example also shows how difficult is to infer the stance without a deep knowledge of the context.
\end{itemize}

After the evaluation of participating systems, the organizers of Semeval-2016 Task 6 
annotated the Stance Datataset for sentiment and target in order to explore the relationship between sentiment and stance \cite{MOHAMMADLREC2016,mohammad-EtAl:2016:SemEval}\footnote{Notice that this is the first publicly available 
Twitter dataset annotated with both stance and sentiment.}. 
In particular, tweets were manually annotated by using two additional labels: \textit{Sentiment} and 
\textit{Opinion Towards}, used to mark the overall sentiment polarity of the tweet and information about the fact that opinion is expressed directly towards the target, respectively:
\begin{itemize}[leftmargin=*]
    \item \textbf{Sentiment}. It can be positive, negative, neutral or none.
    \item \textbf{Opinion\_target}. It can take three different values: (1) if a tweet expresses an opinion about the target; (2) if a tweet expresses an opinion related to an aspect of the target or related to something that is not the target; and (3) if there is not opinion expressed.
\end{itemize}
We decided to exploit such new labels, by enriching our model with corresponding \textbf{labeled-based features}, 
with the aim to experiment with both context and sentiment information provided by human annotators.

\section{Evaluation}

We experimented with a set of tweets belonging to Hillary Clinton and Donald Trump from the Stance Dataset, the Table \ref{distribution_stance} shows the distribution of tweets annotated with stance in the training and the test set for our  targets of interest.

\begin{table}[h]
\centering
\caption{Distribution of stance in training and test set}
\label{distribution_stance}
\begin{tabular}{lcccc|cccc}
\hline
\multirow{2}{*}{Targets} & \multicolumn{4}{c|}{\% Instances in training} & \multicolumn{4}{c}{\% Instances in test} \\ \cline{2-9} 
& Total &  Against & Favor & None & Total & Against & Favor & None\\ \hline
\begin{tabular}[c]{@{}l@{}}Hillary Clinton \end{tabular} 
& 689  & 57.1      & 17.1       & 25.8  & 295     & 58.3 & 15.3 & 26.4      \\ \cline{1-1}
\begin{tabular}[c]{@{}l@{}}Donald Trump  \end{tabular}
& -  & -      & -       & -  & 707     & 42.3 & 20.9 & 36.8        \\ \cline{1-1}
\end{tabular}
\end{table}

We evaluated our approach by using the same measure defined in \cite{mohammad-EtAl:2016:SemEval}  in order to compare our results with those participating in the task.
We trained a Gaussian Naive Bayes classifier \cite{Chan:1979:UFP:892239} implemented in Scikit-learn Python library\footnote{\url{http://scikit-learn.org/}}  to built a model for identifying stance in tweets.

We adopted two experimental setting: a) \textbf{experiment1}. It means to the use of the Sentiment-based, Structural and Context-based features; b) \textbf{experiment2}. It refers to the use of all the features described in Section \ref{ourApproach} including the labeled-based ones. Besides, we experimented using different feature combinations in order to identify which kinds of features could be more relevant for stance detection. 
\begin{table}[h]
\centering
\caption{Best features combination for Hillary Clinton, and the respective results for Donald Trump with experiment1 setting}
\label{table_results_1}
\begin{tabular}{p{5.5cm}ccc|ccc}
\hline
\multirow{2}{*}{Feature set} & \multicolumn{3}{c|}{Hillary Clinton} & \multicolumn{3}{c}{Donald Trump} \\ \cline{2-7}
& F\textsubscript{avg} & F\textsubscript{against} & F\textsubscript{favor}  & F\textsubscript{avg} &F\textsubscript{against} & F\textsubscript{favor}  \\ \hline
\begin{tabular}[c]{@{}l@{}}mention punct\_marks\\ AFINN LIWC HL context\_based\end{tabular}
& 63.75  & 71.95      & 55.56 & 53.46  & 50.29 & 56.63  \\ \cline{1-1}
\begin{tabular}[c]{@{}l@{}}punct\_marks\\ AFINN LIWC HL context\_based \end{tabular}
& 62.70  & 71.47 & 53.93 & 52.76 & 49.61 & 55.91 \\ \cline{1-1}
\begin{tabular}[c]{@{}l@{}}hashtag punct\_marks\\ AFINN LIWC HL DAL context\_based\end{tabular}
& 62.3 & 70.43 & 54.17 & 50.44  & 47.69 & 53.19  \\ \cline{1-1}
\end{tabular}
\end{table}

\begin{table}[h]
\centering
\caption{Feature set for Hillary Clinton and the respective results for Donald Trump with experiment2}
\label{table_results_2}
\begin{tabular}{lccc|ccc}
\hline
\multirow{2}{*}{Feature set} & \multicolumn{3}{c|}{Hillary Clinton} & \multicolumn{3}{c}{Donald Trump} \\ \cline{2-7} 
& F\textsubscript{avg} & F\textsubscript{against} & F\textsubscript{favor}  & F\textsubscript{avg} & F\textsubscript{against} & F\textsubscript{favor}  \\ \hline
\begin{tabular}[c]{@{}l@{}}hashtag  mention\\  context-based labeled-based \end{tabular} 
& 71.21  & 77.17      & 65.26       & 69.59  & 61.99      & 77.19       \\ \cline{1-1}
\begin{tabular}[c]{@{}l@{}}hashtag context-based labeled-based  \end{tabular}
& 71.02  & 76.77      & 65.26       & 70.40  & 62.77      & 78.48       \\ \cline{1-1}
\begin{tabular}[c]{@{}l@{}}hashtag mention\\ LIWC context-based labeled-based \end{tabular}
& 70.98  & 78.23      & 63.73       & 70.20  & 63.06      & 77.35       \\ \cline{1-1}
\end{tabular}
\end{table}
Tables \ref{table_results_1} and \ref{table_results_2} present the best results obtained for Hillary Clinton in the experiment1 and experiment2, respectively. Moreover, those obtained by using the same set of features for Donald Trump are shown.
From the results can be noted that the F1-score in ``against" class is higher than in ``favor". Interestingly, the opposite happens for Donald Trump. 
The results in Table \ref{table_results_2} are higher than those from Table \ref{table_results_1}.
Table \ref{table_results_3} shows the best results for Donald Trump using for both experiment1 and experiment2.

\begin{table}[h]
\caption{Best feature set for Donald Trump using experiment2 and experiment1 setting}
\label{table_results_3}
\begin{tabular}{p{7cm}p{1.5cm}p{1.5cm}p{1.5cm}p{1.5cm}}
\hline
\multirow{2}{*}{Feature set}                                                                   & \multicolumn{3}{c}{Donald Trump}  \\ \cline{2-4} 
                                                                                               & F\textsubscript{avg} & F\textsubscript{against} & F\textsubscript{favor}  \\ \hline
\begin{tabular}[c]{@{}l@{}}* LIWC HL context\_based  labeld\_based\end{tabular} & 74.49  & 69.26      & 79.72       \\ \cline{1-1}
\begin{tabular}[c]{@{}l@{}}mention punc\_marks HL context\_based\end{tabular}              & 55.51  & 50        & 61.02        \\ \cline{1-1}
\end{tabular}
\begin{tablenotes}
  \item The * indicate the use of features belonging exclusively to experiment2.
\end{tablenotes}
\end{table}


\begin{table}[h]
\centering
\caption{Results of task A and B}
\label{ranktable}
\begin{tabular}{lcccc}
\hline
\multicolumn{1}{c}{\multirow{2}{*}{}}     & \multicolumn{2}{c|}{Task A: Hillary Clinton}                       & \multicolumn{2}{c}{Task B: Donald Trump}            \\ \cline{2-5} 
\multicolumn{1}{c}{}                                     & F\textsubscript{avg}               & \multicolumn{1}{c|}{Ranking}           & F\textsubscript{avg}               & Ranking               \\ \hline
experiment1 for Hillary Clinton  & \textbf{63.75}                & \multicolumn{1}{c|}{3}              & 53.46                & 2                    \\ \cline{1-1}
experiment1 for Donald Trump  & 61.25                & \multicolumn{1}{c|}{4}              & \textbf{55.51}                & 2                    \\ \cline{1-1}
experiment2 for Hillary Clinton   & \textbf{71.21}                & \multicolumn{1}{c|}{1}              & 69.59                & 1                    \\ \cline{1-1}
experiment2 for Donald Trump  & 68.29                & \multicolumn{1}{c|}{1}              & \textbf{74.49}                & 1                    \\ \hline
\multicolumn{1}{c}{Systems in the official competition} & \multicolumn{1}{l}{} & \multicolumn{1}{l}{}                & \multicolumn{1}{l}{} & \multicolumn{1}{l}{} \\ \hline
INF-UFRGS                                                & -       & \multicolumn{1}{c|}{-} & 42.32                & 3                    \\ \cline{1-1}
LitisMind                                               & 42.08                & \multicolumn{1}{c|}{17}             & 44.66                & 2                    \\ \cline{1-1}
pkudbblab                                                & 64.41                & \multicolumn{1}{c|}{2}              & \textbf{56.28}               & 1                    \\ \cline{1-1}
PKULCWM                                                  & 62.26                & \multicolumn{1}{c|}{3}              & -       & -      \\ \cline{1-1}
TakeLab                                                  & \textbf{67.12}               & \multicolumn{1}{c|}{1}              & -       & -       \\ \hline
\end{tabular}
\end{table}

As can be noted the context-based features seem to be so relevant for both targets. 
Besides, it is important to highlight that the best result for each target was not achieved by the same set of features.
This is maybe not surprising, if we consider the different political campaign marketing strategies of the two candidates,
which can influence also the communication of candidates' oppositors and supporters, both in terms of language register used and
addressed topics.
For the sake of comparison with the state of the art, we present the results obtained by the three best ranked systems at SemEval-2016 Task 6.
We only include the results concerning to Hillary Clinton and Donald Trump. 
Both the F-measure average and the rank position of each system are included in the Table \ref{ranktable}.
We also show our best results for the two targets using both experimental settings as well as the position in the official ranking in the shared task.

Our approach achieves strongly competitive results.
We ranked in the first position for both Task A and Task B using the experiment2 setting considering Hillary Clinton and Donald Trump results.
For what concerns to the experiment1 we ranked in the third position for Task A and the second one for Task B.
The obtained results outperform the baselines proposed in \cite{mohammad-EtAl:2016:SemEval}\footnote{The authors experimented with n-grams, char-grams and  majority class to establish the baselines for the task.}.
Besides, our outcomes outrank those obtained by submissions from all teams participating in the shared task (both task A and B).
Overall, the results for Hillary Clinton are higher than those for Donald Trump.
This was in someway expected, due to the lack of a training set of tweets concerning the target \textit{Donald Trump}.

\section{Conclusions} 

In this paper we have shown that including context-related information is crucial in order to improve the performance of stance detection systems. Experiments confirms that stance detection is highly dependent on the domain knowledge of the target in hand. 
Our approach relies on the presence of entities related to a target in order to try to extract the opinion expressed towards it.
Besides, our proposal allows to infer the stance in both cases when the target is explicitly mentioned and also when it is not.
The results obtained by exploiting context-related features outperforms those from the best ranked systems in the SemEval-2016 Task 6.

Let us highlight that we are not using either n-grams or any word-based representation, but our approach mainly relies on 
the context of the target in hand. We plan to investigate the performance of our approach in different domains.
Exploiting semantic resources in order to catch additional context information is also an interesting line for future research.
Also user's information and her social network structure could be useful. 
For what concerns to the sentiment-related features, overall results confirm that these kinds of features help 
in identifying the stance towards a particular target. 
We exploited different sentiment-related features, ranging from those extracted from affective resources 
to manually assigned polarity labels. 

A further interesting matter of future work could be explore also the stance w.r.t. different aspects of a political target
entity. This means to perform a sort of aspect-based sentiment analysis in a political domain, e.g., a tweeter can be in favor of 
Hillary for aspects related to ``Health", but not for other aspects.

Finally, we think that it could be also interesting to investigate how to fruitfully combine information about stance and information about the presence of figurative devices in tweets, such as irony and sarcasm \cite{ACM-TOIT-Farias:2016,KBS-Sulis:2016}, since the use of such devices is very frequent in political debates also in social media and detecting irony and sarcasm have been considered as one of the biggest challenges for sentiment analysis.



\section*{Acknowledgments}

The National Council for Science and Technology (CONACyT Mexico) has funded the research work of Delia Irazú Hernández Farías (218109/313683). 
The work of Paolo Rosso has been partially funded by the SomEMBED TIN2015-71147-C2-1-P MINECO research project and by the Generalitat Valenciana under the grant ALMAMATER (PrometeoII/2014/030). The work of Viviana Patti was partially carried out at the Universitat Polit\`ecnica de Val\`encia within the framework of a fellowship of the University of Turin co-funded by Fondazione CRT (World Wide Style Program 2).
\bibliographystyle{splncs}
\bibliography{paper}
\end{document}